\documentclass[11pt,letterpaper]{article}

\usepackage[title]{appendix}

\usepackage{caption}
\usepackage{subcaption}
\usepackage{fullpage}
\usepackage{times}
\usepackage{array}
\usepackage{ragged2e}
\usepackage{multirow}
\usepackage{fancyhdr, amssymb, mathtools}
\usepackage[ruled,vlined]{algorithm2e}
\usepackage[dvipsnames,table]{xcolor}
\usepackage[margin = 1 in]{geometry}
\usepackage{graphicx}
\usepackage{outlines}
\usepackage{amsmath}

\usepackage{graphicx}
\usepackage{dirtytalk}

\usepackage{multirow}
\usepackage{arydshln}

\usepackage[english]{babel} 
\usepackage{blindtext}
\usepackage{amsfonts}
\usepackage{amssymb}
\usepackage{bbm}
\usepackage{mathtools}

\usepackage{hhline}
\usepackage{makecell}

\usepackage{xltabular}
\usepackage{arydshln} 
\usepackage{wrapfig}
\usepackage{enumitem}
\usepackage{floatrow}
\usepackage{xfrac}
\usepackage[font={footnotesize}, labelfont={bf}, labelsep=space, justification=justified, skip=0pt]{caption}
\usepackage[hang,flushmargin]{footmisc} 
\usepackage[colorlinks=true, allcolors=blue]{hyperref}
\usepackage{booktabs}
\usepackage{csquotes}

\definecolor{headcolor}{RGB}{255, 255, 255}
\definecolor{columncolor}{RGB}{255, 255, 255}
\definecolor{modelcolor}{RGB}{255, 255, 255}
\definecolor{optioncolor}{RGB}{255, 255, 255}

\definecolor{codegreen}{rgb}{0,0.6,0}
\definecolor{codegray}{rgb}{0.5,0.5,0.5}
\definecolor{codepurple}{rgb}{0.58,0,0.82}
\definecolor{backcolour}{rgb}{0.95,0.95,0.92}
\usepackage{siunitx}
\usepackage{listings}
\lstdefinestyle{mystyle}{
    backgroundcolor=\color{backcolour},   
    commentstyle=\color{codegreen},
    keywordstyle=\color{magenta},
    numberstyle=\tiny\color{codegray},
    stringstyle=\color{codepurple},
    basicstyle=\ttfamily\footnotesize,
    breakatwhitespace=false,         
    breaklines=true,                 
    captionpos=b,                    
    keepspaces=true,                 
    numbers=left,                    
    numbersep=5pt,                  
    showspaces=false,                
    showstringspaces=false,
    showtabs=false,                  
    tabsize=2
}
\usepackage{setspace}
\usepackage[numbers,sort&compress]{natbib}
\usepackage{cleveref} 
\AtBeginEnvironment{appendices}{\crefalias{section}{appendix}}
\newcommand{\cmt}[1]{} 

\newcommand{\corrsym}{\protect\footnotemark[1]}
    \makeatletter
\def\@fnsymbol#1{\ensuremath{\ifcase#1\or \dagger\or *\or \ddagger\or
   \mathsection\or \mathparagraph\or \|\or **\or \dagger\dagger
   \or \ddagger\ddagger \else\@ctrerr\fi}}
    \makeatother

\usepackage{titling}
\thanksheadextra{}{}
\thanksheadextra{}{} 
\usepackage{lipsum} 

\title{On the Uncertainty Quantification Ability of Tabular Foundation Models}
\date{\vspace{-5ex}}
\usepackage{authblk}
\author[1]{Tyler R. Johnson}
\author[1]{Kian Ben-Jacob}
\author[1]{Nima Negarandeh}
\author[1]{Oriol Vendrell-Gallart}
\author[1,2]{Ramin Bostanabad\corrsym}  
\affil[1]{Department of Mechanical and Aerospace Engineering, University of California, Irvine}
\affil[2]{Department of Civil and Environmental Engineering, University of California, Irvine}

\begin{document}

    \pagenumbering{arabic}
    \sloppy
    \maketitle
    
    \begingroup
    \renewcommand{\thefootnote}{}
    \footnotetext{\textsuperscript{†}Corresponding Author: Raminb@uci.edu}
    \endgroup
    \section*{Abstract}
\looseness-1 Foundation models (FMs) have achieved substantial success in generalizing across tasks without problem-specific training or fine-tuning. However, many critical applications in mechanics and computational science require not only accurate predictions but also reliable uncertainty quantification (UQ). 
Herein we investigate the UQ capabilities of tabular FMs in regression tasks through a comprehensive empirical study comparing Tabular Prior-Data Fitted Networks (TabPFN) against Gaussian processes (GPs). We systematically evaluate these two methods across a host of regression problems with varying complexity, dataset sizes, and input dimensionalities. 
We use a default setting to build all the GPs and for a fair comparison against TabPFN v2.5.
Our findings highlight an important trade-off between explicit and learned priors: while TabPFN achieves highly competitive performance for complex, high-dimensional problems with sufficient data, GPs often provide superior predictive accuracy and UQ in data-scarce settings. Moreover, when the chosen kernel constitutes a good prior for the underlying function, GP performance can substantially exceed that of TabPFN. Our results can be reproduced from \hyperlink{https://github.com/kianswarehouse/GPvsPFN}{https://github.com/kianswarehouse/GPvsPFN}.

\noindent \textbf{Keywords:} Gaussian Processes;Tabular Foundational Models; Uncertainty Quantification. 

    \section{Introduction} \label{sec introduction}

Surrogates are increasingly employed to replace or augment expensive simulations. However, selecting the appropriate surrogate type, e.g., a neural network (NN) or a Gaussian process (GP), and training it remains a non-trivial task, particularly when training data is noisy, scarce, or high-dimensional. Consequently, there is growing interest in the engineering and scientific communities in developing probabilistic surrogates, or \textit{emulators}, capable of uncertainty quantification (UQ).

Over the past few decades many emulation techniques have been developed and a non-exhaustive list of the most common ones includes Bayesian NNs \cite{RN1635}, Gaussian processes (GPs) \cite{RN332}, polynomial chaos expansions \cite{RN466}, and ensemble- or dropout- based approaches \cite{RN1074,RN1055}. These methods provide theoretically strong backbones for UQ and have seen tremendous success in various applications. A feature of all these ``traditional'' methods is that they rely on a training or sampling stage. Foundation models (FMs) have emerged as powerful alternatives that aim to dispense with this feature.

FMs are a class of machine learning (ML) models that are expected to transfer across tasks with minimal tuning. Beyond providing high accuracy and scalability across different modalities, recent work has shown that transformer-like architectures \cite{lin2021surveytransformers} can approximate Bayesian posterior predictive quantities by learning from samples drawn from a specified prior over datasets. This approximation capability is especially relevant for UQ as it enables probabilistic predictions for new tasks without per-task retraining. Building on this idea, tabular foundation models such as TabPFN \cite{Hollmann2025NatureTabPFN} have emerged as powerful FMs for probabilistically learning from small-to-medium tabular datasets.

\begin{figure*}[!t]
    \centering
    \includegraphics[trim={0 30 0 0}, width=0.9\linewidth]{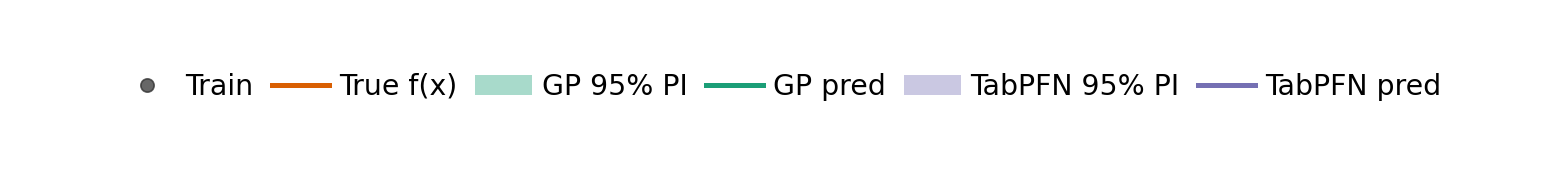}
    \begin{subfigure}{0.19\textwidth}
        \centering
        \includegraphics[width=1.0\linewidth]{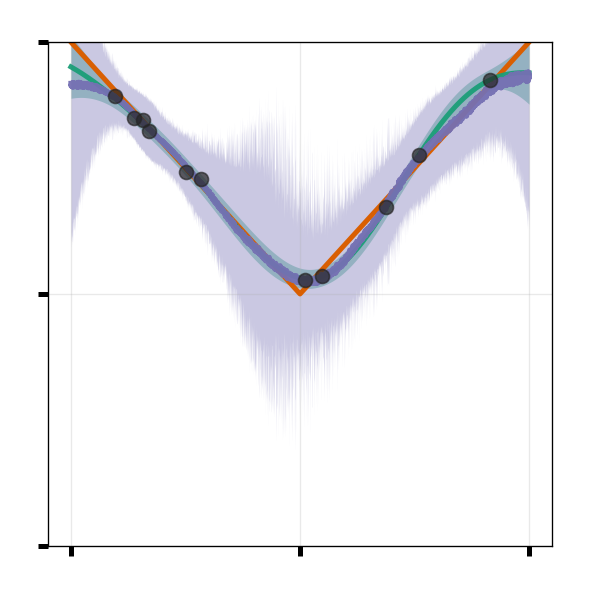}
        \captionsetup{font=scriptsize}
        \caption{$\left|x\right|$}
        \label{subfig abs x}
    \end{subfigure}
    \hfill
    \begin{subfigure}{0.19\textwidth}
        \centering
        \includegraphics[width=1.0\linewidth]{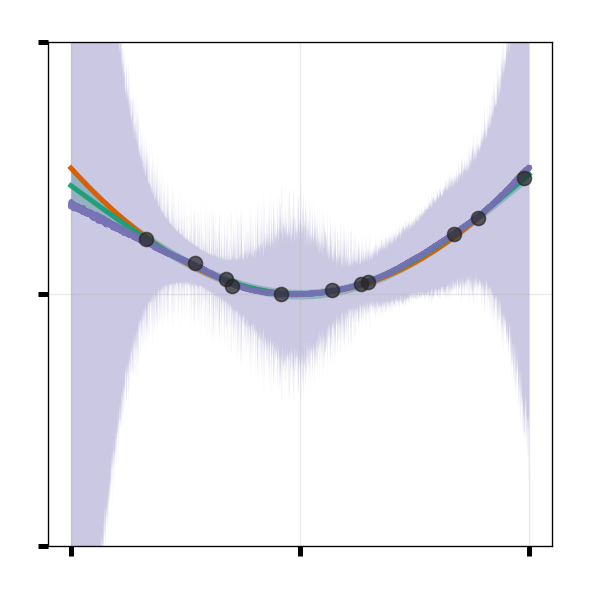}
        \captionsetup{font=scriptsize}
        \caption{$x^2$}
        \label{subfig x squared}
    \end{subfigure}
    \hfill
    \begin{subfigure}{0.19\textwidth}
        \centering
        \includegraphics[width=1.0\linewidth]{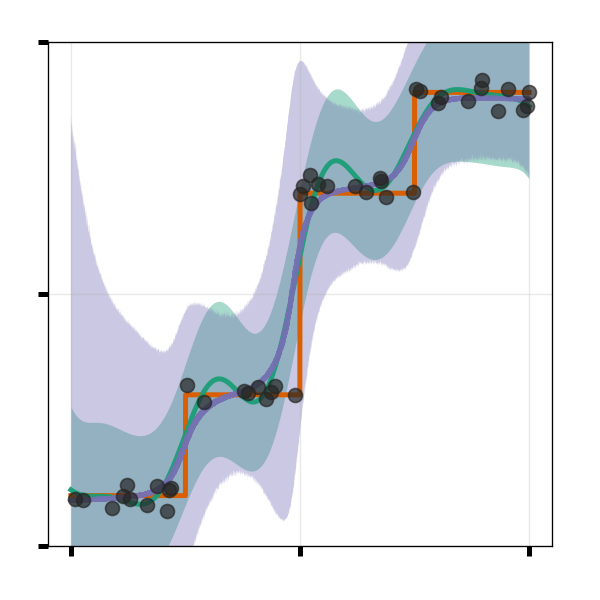}
        \captionsetup{font=scriptsize}
        \caption{Step Function}
        \label{subfig step}
    \end{subfigure}
    \hfill
    \begin{subfigure}{0.19\textwidth}
        \centering
        \includegraphics[width=1.0\linewidth]{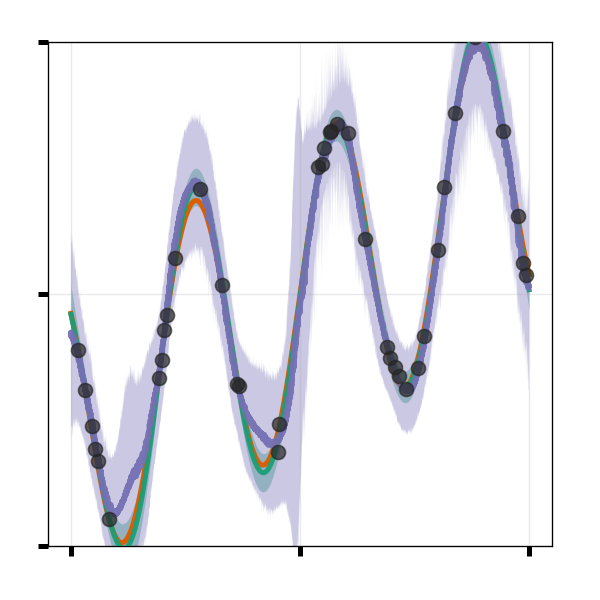}
        \captionsetup{font=scriptsize}
        \caption{$0.3$ sin$(6.5{\pi}x) + 0.5x$}
        \label{subfig sin x plus x}
    \end{subfigure}
    \hfill
    \begin{subfigure}{0.19\textwidth}
        \centering
        \includegraphics[width=1.0\linewidth]{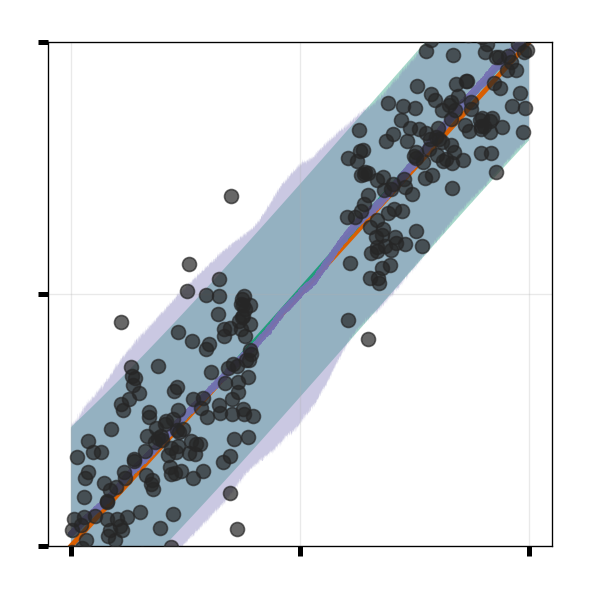}
        \captionsetup{font=scriptsize}
        \caption{Gappy Data}
        \label{subfig linear homoscedastic}
    \end{subfigure}
    \caption{\textbf{1D Examples:} The functions are all adopted from \cite{Hollmann2025NatureTabPFN}. Each plot has the same x and y axes, both spanning $[-0.5, 0.5]$. All the GPs are trained on CPU using the defaults of \cite{RN2079} (e.g., constant mean and a Gaussian kernel). TabPFN v2.5 is run on GPU without fine-tuning.}
    \label{fig 1d analytic ex}
\end{figure*}

These developments motivate a practical and timely study to assess whether FMs such as TabPFN deliver uncertainty estimates that are as reliable as those produced by classical methods such as GPs. Motivated by this need, in this work we empirically evaluate the UQ ability of TabPFN in regression problems by comparing it to GP baselines across different problems, noise levels, dataset sizes, and input dimensionalities, see Figure 1 for some 1D examples. We focus on reproducible, practitioner-relevant usage by evaluating default implementations throughout, and we measure both predictive accuracy and UQ quality. 
Our primary contributions are:
\begin{itemize}
    \item We recommend a principled set of defaults for building GPs to increase their performance across different applications.
    \item We open-source our contributions via GP+ \cite{RN2079} and compare its performance to TabPFN v2.5 using three metrics that measure point prediction accuracy (relative root mean squared error or RRMSE), predictive distribution quality (relative mean negatively oriented interval score or NIS \cite{RN284}), and computational costs.
    \item We discuss limitations of GPs and TabPFN for predictive modeling and UQ--providing a practical guideline for emulator selection.
\end{itemize}

Below, we first provide relevant technical background and then elaborate on our comparison methodology. 
Our findings are summarized in the Results Section which is followed by the concluding remarks. 
    \section{Technical Background}
\subsection{TabPFN as a Tabular Foundation Model} \label{sec tabpfn background}

TabPFN is built on the Prior-Data Fitted Network (PFN) framework, which trains a neural model to approximate the Bayesian posterior predictive distribution induced by a user-defined prior over supervised learning tasks. In this setup, one first specifies a prior over datasets by defining how to sample synthetic tasks. Then, training repeatedly draws datasets from this prior and optimizes the model to predict held-out samples. The PFN training objective is designed such that minimizing it makes the model's predictive distribution close to the true posterior predictive under the specified prior, yielding a Bayesian interpretation: after offline training, the model can output posterior predictive distributions for new datasets through conditioning on the provided context.

Under this view, TabPFN's uncertainty behavior is tightly linked to the designed prior since the model is trained to return predictive distributions consistent with the posterior predictive implied by the synthetic data-generating process. The original TabPFN constructs synthetic tabular tasks from a mixture of generative mechanisms, including causal-style generators and Bayesian NNs, to expose the model to diverse feature-target relationships.

Later versions refine the synthetic-task distribution to better reflect practical tabular issues that affect both accuracy and uncertainty (i.e., incorporating corruptions such as missing values, uninformative features, and outliers during pretraining) \cite{Hollmann2025NatureTabPFN}. TabPFN v2.0 introduces a randomized feature-token mechanism which handles heterogeneous feature spaces, supporting more reliable transfer of predictions across datasets with differing feature semantics. TabPFN v2.5 further advances the architecture by refining the transformer design and training procedure, resulting in improved predictive accuracy and uncertainty estimates across a broader range of tabular datasets.

Architecturally, TabPFN treats an entire dataset as the input object and uses attention to condition each query prediction on the full labeled context. This dataset-level conditioning means runtime comparisons must distinguish end-to-end inference (conditioning plus prediction) from ``traditional'' post-fit prediction time in conventional fitted models; standard baselines can be faster once fitted, because TabPFN effectively performs dataset conditioning and prediction jointly \cite{Hollmann2025NatureTabPFN}.

The computational cost of transformer attention scales quadratically in the number of rows processed in a forward pass \cite{lin2021surveytransformers}. For TabPFN-style in-context inference, if $N_c$ is the number of labeled context (training) rows and $N_q$ is the number of query rows, the computational cost scales as:
\begin{equation}
    \mathrm{Cost} = \mathcal{O}\!\left(N_c(N_c+N_q)\right),
\end{equation}
since query rows attend only to training rows and not to each other \cite{Hollmann2025NatureTabPFN}. Recent versions of TabPFN mitigate this by caching the training-context state, amortizing the $\mathcal{O}(N_c^2)$ cost across repeated queries against a fixed training set.

\subsection{Gaussian Processes} \label{sec gp background}
GPs provide a natural Bayesian framework for function approximation by placing a prior on functions and then updating that prior based on data \cite{RN332}. They are completely characterized by their mean function $\textit{m}(\boldsymbol{x}; \boldsymbol{\theta})$ and kernel or covariance function $\textit{k}(\boldsymbol{x}, \boldsymbol{x'}; \boldsymbol{\beta})$ where $\boldsymbol{\theta}$ and $\boldsymbol{\beta}$ are hyperparameters that are conventionally estimated based on a training dataset. 



The optimal choice of mean and covariance functions depends on the problem but it is common practice to use a zero or constant mean and a stationary kernel such as the Gaussian or Mat\'ern \cite{RN2089}. The power exponential (PE) kernel is another useful stationary kernel defined as ($\boldsymbol{\beta} = \{s, \omega_i, p\}$ are the hyperparameters):
\begin{equation}
    k(\mathbf{x}, \mathbf{x}'; \boldsymbol{\beta}) = 10^s \exp\left(-\sum_{i=1}^{D_x} 10^{\omega_i}|x_i - x_i'|^p\right)
    \label{eq pe kernel}
\end{equation}
which leverages automatic relevance determination and is equivalent to the Gaussian kernel for $p=2$. 

In practice, regression or interpolation with GPs requires two major steps: (1) selecting the mean and covariance functions, and (2) using the training data for hyperparameter optimization which is typically done via maximum likelihood estimation (MLE). FMs such as TabPFN require neither of these steps and hence are very attractive alternatives since the performance of the GP can be sensitive to the selected mean or covariance functions as well as the training process. 


GPs have additional limitations such as scalability since the cost of repeatedly inverting the covariance matrix during training scales as $O(N^3)$ where $N$ is the number of samples. Their performance also degrades in very high dimensions since selecting an appropriate kernel and effectively estimating its hyperparameters becomes very challenging. Over the past two decades, many effective methods have been developed that address these limitations of GPs \cite{RN413, RN414, RN893, RN2110, RN1875, RN1055, RN1901} but herein our focus is on ``vanilla'' GPs that do not leverage any of these advancements. 

We train the GPs following \cite{RN2079} and introduce effective defaults on initialization, formulations, and optimization to produce high-performing emulators across multiple applications. We refer the reader to the GitHub page for details and merely provide an example here by noting that the kernel in \Cref{eq pe kernel} is commonly formulated as:
\begin{equation}
    k(\mathbf{x}, \mathbf{x}'; \boldsymbol{\beta}) = \sigma^2 \exp\left(-\sum_{i=1}^{D_x} \left(\frac{|x_i - x_i'|}{l_i} \right)^p \right),
    \label{eq pe kernel typically}
\end{equation}
whose hyperparameters are frequently more challenging to optimize than those in \Cref{eq pe kernel}.
    \section{Comparison Methodology} \label{sec method}

We design a set of controlled experiments to compare the emulation power of GPs and TabPFN v2.5 on a host of regression problems. Five of these problems (Wing Weight, Ackley, Dixon-Price, Griewank, and Rosenbrock) are taken from \cite{simulationlib} while the Buckling function is adopted from \cite{OUNE2021114128} to include categorical inputs in our studies. For these six problems, the input dimensionality $D_x$ ranges between $4$ and $40$ while the output dimensionality is $1$ in all problems since TabPFN only accommodates single-response datasets. To include the effect of data size and noise variance on our analyses, for each problem we consider two training dataset sizes, $N = 10D_x$ and $N=40D_x$, and two noise levels, $\varepsilon \sim \mathcal{N}(0, (0.005\times c)^2)$ and $\varepsilon \sim \mathcal{N}(0, (0.05\times c)^2)$, where $c$ is the standard deviation of the target's test data, i.e., $c$ is a problem-dependent constant that adjusts the noise variance based on the problem. 

To study the effect of misspecified likelihoods, we also consider non-Gaussian noise in our analytic examples and study two real-world regression benchmarks, namely, Elevators ($D_x=18$) and Pumadyn ($D_x=32$). 
For evaluation, for each analytic example we generate a single large test dataset of size $N_{test}=5000$ via a Sobol sequence whose seed number is chosen a priori to ensure TabPFN and GP are compared on the same data. For the two real-world benchmarks, random subsets of the data are used for training and we reserve a single large dataset, $N_{test}=3000$, for testing (if the benchmark has insufficient data to train on 20 independent datasets, the subsets were allowed to have some overlap). The median values of our metrics across 20 random sets are reported (see GitHub for statistics).



Let \(y_i\) and \(\hat{y}_i\) denote the true and predicted values for \(i = 1, \ldots, N_{test}\). RRMSE measures point prediction accuracy and is calculated as:
\begin{equation}
  \mathrm{RRMSE}
  = \frac{1}{c} \sqrt{\frac{1}{N_{test}} \sum_{i=1}^{N_{test}} (y_i - \hat{y}_i)^2}.
  \label{eq rrmse}
\end{equation}
where $c$ is a problem-dependent scalar (specifically, the response standard deviation in the test data) that ensures RRMSEs are comparable across different problems. 
To evaluate predictive distribution quality we use NIS. The negatively oriented interval score for a central \((1-\alpha)\) prediction interval \([L_i, U_i]\) is:
\begin{equation*}
  \mathrm{S}_i
  = \underbrace{(U_i - L_i)}_{a_i}
  + \underbrace{\frac{(L_i - y_i) \mathbf{1}[y_i < L_i]}{\alpha / 2} + \frac{(y_i - U_i) \mathbf{1}[y_i > U_i]}{\alpha / 2}}_{b_i},
\end{equation*}
where \(\mathbf{1}[\cdot]\) is the indicator function, \(\alpha \in (0,1)\) (e.g.\ \(\alpha = 0.05\) for a 95\% interval), $a_i$ denotes the interval width, and $b_i$ is the penalty for falling below \(L_i\) or above \(U_i\). NIS is then calculated as:
\begin{equation}
    \mathrm{NIS}
    = \frac{1}{c}\, \frac{1}{N_{test}} \sum_{i=1}^{N_{test}} a_i + b_i = A + B,
    \label{eq nis}
\end{equation}
where $c$ is the same scalar as in \Cref{eq rrmse}.

For TabPFN, we construct a central 95\% predictive interval by taking the 2.5th and 97.5th percentiles of the predictive distribution at each test input $x_i$:
\[
[L_i, U_i] = \big[Q_{0.025}(y\mid x_i),\; Q_{0.975}(y\mid x_i)\big].
\]
Because this interval is quantile-based, it need not be symmetric about the predictive mean.
For GPs, we report a Gaussian 95\% interval using the predictive mean $\mu_i$ and standard deviation $\sigma_i$.

We emphasize that, unlike TabPFN, GPs require training which typically requires some fine-tuning on the selection of the kernel, hyperparameter priors, and optimization settings. However, we refrain from this fine-tuning and train the GPs using the \textit{defaults} of our GP+ package \cite{RN2079}. Obviously, this choice indicates that our GPs will \textit{not} be the best (or maybe even close to the best) GPs one can build, but we have made this choice since the competing method (i.e., TabPFN) does not require the user to spend any time on training. Additionally, TabPFN does not require any preprocessing. So, we primarily rely on the default preprocessing of GP+ which scales each input to $[0, 1]$ and standardizes the output vector based on its mean and variance.


Training and prediction are done on CPU for GPs but since TabPFN is a transformer-based model we use it on GPU (TabPFN provides almost identical performance metrics on CPU but at a significantly higher wall clock time, see \Cref{tab results}). The hardware used in these experiments is the 11th Gen Intel{\textregistered} Core{\texttrademark} i7-11700K @ 3.6GHz and the NVIDIA GeForce RTX 3060 (12GB) for CPU and GPU, respectively. 

    \section{Results and Discussions} \label{sec results}

The main results of our studies are summarized in \Cref{tab results} and \Cref{fig results}, where the latter only visualizes a subset of \Cref{tab results} due to space constraints. Note that while the medians across 20 independent runs are representative, individual runs may show different relative rankings between the two models on either RRMSE or NIS (see GitHub for details). We analyze these results below in terms of accuracy, UQ, cost, and ease-of-use of the models. We note that we also experimented with adjusting multiple TabPFN configuration settings, but observed no meaningful improvement over the defaults, suggesting that out-of-the-box TabPFN already performs near its ceiling on these benchmarks.

\begin{table*}[!b]
    \centering
    {\extracolsep{-10pt}
    \setlength{\tabcolsep}{5pt}
    \renewcommand{\arraystretch}{1.15}
    \tiny
    \caption{\textbf{Summary of results:} RRMSE, NIS, and wall-clock time are reported (medians across 20 runs, see GitHub for statistics). Best numbers are in bold font and noise refers to either $\varepsilon \sim \mathcal{N}(0, (0.005\times c)^2)$ or $\varepsilon \sim \mathcal{N}(0, (0.05\times c)^2)$ where $c$ is a problem-dependent constant that adjusts the noise variance based on the problem (see GitHub for non-Gaussian noise). All GPs use default GP+ settings.
    TabPFN runs on GPU (CPU time in parentheses) and the compute time is almost entirely for inference. GP runs on CPU with the majority of its compute time spent during training.}
    \begin{tabular}{>{\centering\arraybackslash}p{1.6cm}>{\centering\arraybackslash}p{0.5cm}>{\centering\arraybackslash}p{0.3cm}>{\centering\arraybackslash}p{0.9cm}|ccc|ccc}
    \hline
    \multicolumn{4}{c|}{} & \multicolumn{3}{c|}{\textbf{GP+}} & \multicolumn{3}{c}{\textbf{TabPFN v2.5}} \\
    \cline{5-10} 
    \textbf{Problem} & \textbf{N} & \textbf{D$_x$} & \textbf{Noise} & \textbf{RRMSE} & \textbf{NIS ($A$ $+$ $B$)} & \textbf{CPU (s)} & \textbf{RRMSE} & \textbf{NIS ($A$ $+$ $B$)} & \textbf{GPU (CPU) (s)} \\
    \hline
    \textbf{Wing Weight} & 10D$_x$ & 10 & 0.005 & \textbf{0.016} & \textbf{0.073} (0.050 + 0.023) & \textbf{1.07} & 0.033 & 0.167 (0.164 + 0.004) & 1.36 (21.98) \\
     & 10D$_x$ & 10 & 0.05 & \textbf{0.068} & 0.331 (0.252 + 0.079) & \textbf{1.11} & 0.070 & \textbf{0.327} (0.294 + 0.034) & 1.38 (21.00) \\
     & 40D$_x$ & 10 & 0.005 & \textbf{0.007} & \textbf{0.033} (0.026 + 0.007) & 7.44 & 0.013 & 0.082 (0.082 + 0.000) & \textbf{1.50} (24.55) \\
     & 40D$_x$ & 10 & 0.05 & \textbf{0.055} & \textbf{0.257} (0.212 + 0.045) & 7.90 & 0.057 & 0.269 (0.224 + 0.045) & \textbf{1.45} (24.54) \\
    \hline
    \textbf{Buckling} & 10D$_x$ & 4 & 0.005 & \textbf{0.174} & 1.567 (0.044 + 1.523) & 9.64 & 0.419 & \textbf{0.656} (0.655 + 0.001) & \textbf{1.10} (13.37) \\
     & 10D$_x$ & 4 & 0.05 & \textbf{0.304} & 3.384 (0.203 + 3.181) & 4.97 & 0.323 & \textbf{1.035} (0.997 + 0.037) & \textbf{1.09} (13.24) \\
     & 40D$_x$ & 4 & 0.005 & \textbf{0.016} & \textbf{0.056} (0.025 + 0.031) & 11.88 & 0.097 & 0.155 (0.153 + 0.002) & \textbf{1.06} (14.18) \\
     & 40D$_x$ & 4 & 0.05 & \textbf{0.078} & \textbf{0.360} (0.215 + 0.145) & 6.42 & 0.231 & 0.469 (0.430 + 0.039) & \textbf{1.07} (13.93) \\
    \hline
    \textbf{Ackley} & 10D$_x$ & 10 & 0.005 & \textbf{0.301} & \textbf{1.569} (1.297 + 0.271) & \textbf{1.16} & 0.378 & 1.887 (1.787 + 0.100) & 1.38 (21.11) \\
     & 10D$_x$ & 10 & 0.05 & \textbf{0.310} & \textbf{1.616} (1.296 + 0.321) & \textbf{1.13} & 0.383 & 1.938 (1.820 + 0.117) & 1.39 (21.49) \\
     & 40D$_x$ & 10 & 0.005 & 0.213 & 1.073 (0.881 + 0.192) & 7.61 & \textbf{0.210} & \textbf{0.993} (0.791 + 0.202) & \textbf{1.48} (25.32) \\
     & 40D$_x$ & 10 & 0.05 & 0.220 & 1.112 (0.920 + 0.192) & 7.88 & \textbf{0.217} & \textbf{1.036} (0.818 + 0.218) & \textbf{1.48} (23.62) \\
     & 10D$_x$ & 20 & 0.005 & \textbf{0.316} & \textbf{1.614} (1.231 + 0.383) & 3.37 & 0.385 & 1.830 (1.709 + 0.121) & \textbf{1.96} (32.07) \\
     & 10D$_x$ & 20 & 0.05 & \textbf{0.321} & \textbf{1.621} (1.234 + 0.387) & 3.42 & 0.390 & 1.866 (1.745 + 0.121) & \textbf{1.97} (31.44) \\
     & 40D$_x$ & 20 & 0.005 & 0.217 & 1.069 (0.901 + 0.168) & 59.54 & \textbf{0.214} & \textbf{0.995} (0.844 + 0.151) & \textbf{2.35} (45.30) \\
     & 40D$_x$ & 20 & 0.05 & 0.225 & 1.096 (0.921 + 0.175) & 60.42 & \textbf{0.222} & \textbf{1.037} (0.877 + 0.160) & \textbf{2.36} (45.18) \\
     & 10D$_x$ & 40 & 0.005 & \textbf{0.298} & \textbf{1.495} (1.265 + 0.230) & 18.00 & 0.416 & 1.962 (1.794 + 0.168) & \textbf{3.20} (62.85) \\
     & 10D$_x$ & 40 & 0.05 & \textbf{0.307} & \textbf{1.531} (1.281 + 0.250) & 17.91 & 0.421 & 1.987 (1.810 + 0.177) & \textbf{3.20} (62.99) \\
     & 40D$_x$ & 40 & 0.005 & 0.243 & 1.182 (0.922 + 0.260) & 467.44 & \textbf{0.230} & \textbf{1.114} (1.003 + 0.111) & \textbf{4.83} (119.17) \\
     & 40D$_x$ & 40 & 0.05 & 0.251 & 1.202 (0.944 + 0.259) & 479.80 & \textbf{0.237} & \textbf{1.138} (1.026 + 0.111) & \textbf{4.85} (118.15) \\
    \hline
    \textbf{Dixon-Price} & 10D$_x$ & 10 & 0.005 & 0.442 & 2.037 (1.539 + 0.498) & \textbf{1.24} & \textbf{0.427} & \textbf{2.010} (1.854 + 0.157) & 1.42 (18.84) \\
     & 10D$_x$ & 10 & 0.05 & 0.453 & 2.085 (1.521 + 0.564) & \textbf{1.25} & \textbf{0.431} & \textbf{2.057} (1.884 + 0.173) & 1.43 (19.04) \\
     & 40D$_x$ & 10 & 0.005 & 0.279 & 1.287 (1.049 + 0.237) & 11.46 & \textbf{0.162} & \textbf{0.840} (0.790 + 0.050) & \textbf{1.53} (21.37) \\
     & 40D$_x$ & 10 & 0.05 & 0.288 & 1.325 (1.080 + 0.245) & 11.40 & \textbf{0.177} & \textbf{0.873} (0.815 + 0.059) & \textbf{1.51} (21.79) \\
     & 10D$_x$ & 20 & 0.005 & 0.474 & 2.286 (1.553 + 0.733) & 4.78 & \textbf{0.469} & \textbf{2.159} (1.911 + 0.247) & \textbf{1.96} (32.29) \\
     & 10D$_x$ & 20 & 0.05 & \textbf{0.466} & 2.227 (1.569 + 0.658) & 4.45 & 0.471 & \textbf{2.172} (1.915 + 0.257) & \textbf{1.97} (32.73) \\
     & 40D$_x$ & 20 & 0.005 & 0.334 & 1.558 (1.300 + 0.258) & 99.93 & \textbf{0.195} & \textbf{0.977} (0.910 + 0.066) & \textbf{2.37} (44.96) \\
     & 40D$_x$ & 20 & 0.05 & 0.339 & 1.561 (1.329 + 0.233) & 99.77 & \textbf{0.203} & \textbf{1.006} (0.930 + 0.076) & \textbf{2.36} (46.18) \\
     & 10D$_x$ & 40 & 0.005 & \textbf{0.455} & \textbf{2.183} (1.618 + 0.565) & 31.05 & 0.480 & 2.327 (2.009 + 0.318) & \textbf{3.20} (62.67) \\
     & 10D$_x$ & 40 & 0.05 & \textbf{0.470} & \textbf{2.240} (1.635 + 0.605) & 32.35 & 0.482 & 2.330 (2.015 + 0.315) & \textbf{3.31} (62.69) \\
     & 40D$_x$ & 40 & 0.005 & 0.431 & 2.086 (1.481 + 0.605) & 867.45 & \textbf{0.231} & \textbf{1.141} (1.066 + 0.075) & \textbf{4.80} (117.45) \\
     & 40D$_x$ & 40 & 0.05 & 0.433 & 2.040 (1.491 + 0.549) & 874.93 & \textbf{0.237} & \textbf{1.163} (1.086 + 0.078) & \textbf{4.85} (119.56) \\
    \hline
    \textbf{Griewank} & 10D$_x$ & 10 & 0.005 & \textbf{0.011} & \textbf{0.111} (0.111 + 0.000) & \textbf{1.22} & 0.512 & 2.460 (2.270 + 0.190) & 1.37 (18.52) \\
     & 10D$_x$ & 10 & 0.05 & \textbf{0.109} & \textbf{0.529} (0.482 + 0.047) & \textbf{1.13} & 0.524 & 2.490 (2.281 + 0.209) & 1.37 (19.48) \\
     & 40D$_x$ & 10 & 0.005 & \textbf{0.007} & \textbf{0.033} (0.028 + 0.005) & 12.01 & 0.100 & 0.469 (0.444 + 0.024) & \textbf{1.45} (21.11) \\
     & 40D$_x$ & 10 & 0.05 & \textbf{0.056} & \textbf{0.262} (0.219 + 0.042) & 11.68 & 0.115 & 0.535 (0.479 + 0.055) & \textbf{1.44} (21.67) \\
     & 10D$_x$ & 20 & 0.005 & \textbf{0.139} & \textbf{1.009} (1.009 + 0.000) & 3.77 & 0.515 & 2.426 (2.175 + 0.251) & \textbf{1.96} (32.05) \\
     & 10D$_x$ & 20 & 0.05 & \textbf{0.174} & \textbf{1.065} (1.060 + 0.005) & 3.80 & 0.516 & 2.436 (2.176 + 0.261) & \textbf{1.95} (32.14) \\
     & 40D$_x$ & 20 & 0.005 & \textbf{0.008} & \textbf{0.040} (0.035 + 0.005) & 88.19 & 0.140 & 0.644 (0.600 + 0.044) & \textbf{2.37} (44.84) \\
     & 40D$_x$ & 20 & 0.05 & \textbf{0.061} & \textbf{0.282} (0.242 + 0.040) & 91.18 & 0.149 & 0.691 (0.629 + 0.062) & \textbf{2.37} (45.61) \\
     & 10D$_x$ & 40 & 0.005 & \textbf{0.255} & \textbf{1.552} (1.543 + 0.009) & 23.33 & 0.548 & 2.549 (2.206 + 0.343) & \textbf{3.19} (62.62) \\
     & 10D$_x$ & 40 & 0.05 & \textbf{0.257} & \textbf{1.562} (1.553 + 0.009) & 23.07 & 0.550 & 2.585 (2.224 + 0.361) & \textbf{3.19} (62.97) \\
     & 40D$_x$ & 40 & 0.005 & \textbf{0.008} & \textbf{0.047} (0.047 + 0.000) & 824.65 & 0.173 & 0.970 (0.901 + 0.068) & \textbf{4.82} (118.70) \\
     & 40D$_x$ & 40 & 0.05 & \textbf{0.074} & \textbf{0.347} (0.299 + 0.048) & 837.12 & 0.181 & 1.007 (0.929 + 0.077) & \textbf{4.83} (118.21) \\
    \hline
    \textbf{Rosenbrock} & 10D$_x$ & 10 & 0.005 & 0.459 & 2.206 (1.622 + 0.584) & \textbf{1.11} & \textbf{0.379} & \textbf{2.084} (2.015 + 0.069) & 1.39 (19.12) \\
     & 10D$_x$ & 10 & 0.05 & 0.462 & 2.285 (1.655 + 0.630) & \textbf{1.17} & \textbf{0.386} & \textbf{2.079} (2.010 + 0.069) & 1.39 (18.61) \\
     & 40D$_x$ & 10 & 0.005 & 0.337 & 1.588 (1.194 + 0.394) & 8.70 & \textbf{0.105} & \textbf{0.500} (0.467 + 0.033) & \textbf{1.49} (21.92) \\
     & 40D$_x$ & 10 & 0.05 & 0.340 & 1.617 (1.226 + 0.391) & 8.74 & \textbf{0.122} & \textbf{0.567} (0.517 + 0.051) & \textbf{1.50} (22.94) \\
     & 10D$_x$ & 20 & 0.005 & 0.490 & 2.435 (1.643 + 0.792) & 3.99 & \textbf{0.468} & \textbf{2.289} (2.157 + 0.132) & \textbf{2.03} (31.58) \\
     & 10D$_x$ & 20 & 0.05 & 0.492 & 2.395 (1.668 + 0.727) & 3.93 & \textbf{0.475} & \textbf{2.302} (2.163 + 0.139) & \textbf{1.92} (32.09) \\
     & 40D$_x$ & 20 & 0.005 & 0.373 & 1.752 (1.428 + 0.324) & 70.29 & \textbf{0.157} & \textbf{0.764} (0.723 + 0.041) & \textbf{2.34} (54.00) \\
     & 40D$_x$ & 20 & 0.05 & 0.377 & 1.774 (1.443 + 0.331) & 66.62 & \textbf{0.165} & \textbf{0.803} (0.749 + 0.054) & \textbf{2.35} (48.69) \\
     & 10D$_x$ & 40 & 0.005 & \textbf{0.481} & \textbf{2.281} (1.683 + 0.598) & 21.93 & 0.501 & 2.371 (2.123 + 0.248) & \textbf{3.19} (62.61) \\
     & 10D$_x$ & 40 & 0.05 & \textbf{0.484} & \textbf{2.297} (1.685 + 0.612) & 21.65 & 0.503 & 2.406 (2.152 + 0.255) & \textbf{3.19} (62.87) \\
     & 40D$_x$ & 40 & 0.005 & 0.422 & 1.997 (1.492 + 0.505) & 592.45 & \textbf{0.215} & \textbf{1.071} (0.991 + 0.081) & \textbf{4.83} (118.52) \\
     & 40D$_x$ & 40 & 0.05 & 0.427 & 2.028 (1.502 + 0.525) & 578.19 & \textbf{0.222} & \textbf{1.099} (1.012 + 0.086) & \textbf{4.82} (118.95) \\
    \hline
    \end{tabular}
    \label{tab results}
    }
\end{table*}

\subsection{Accuracy and Uncertainty Quantification}

Our findings suggest that neither of the methods outperforms the alternative consistently. The overall trends are also consistent in that as dataset size is increased or noise variance is reduced their performance improves and the relative rankings are mostly unchanged.

\begin{figure*}[!b]
    \centering
    \captionsetup{font=footnotesize}
    \begin{minipage}[b]{0.48\textwidth}
        \centering
        \includegraphics[width=\textwidth]{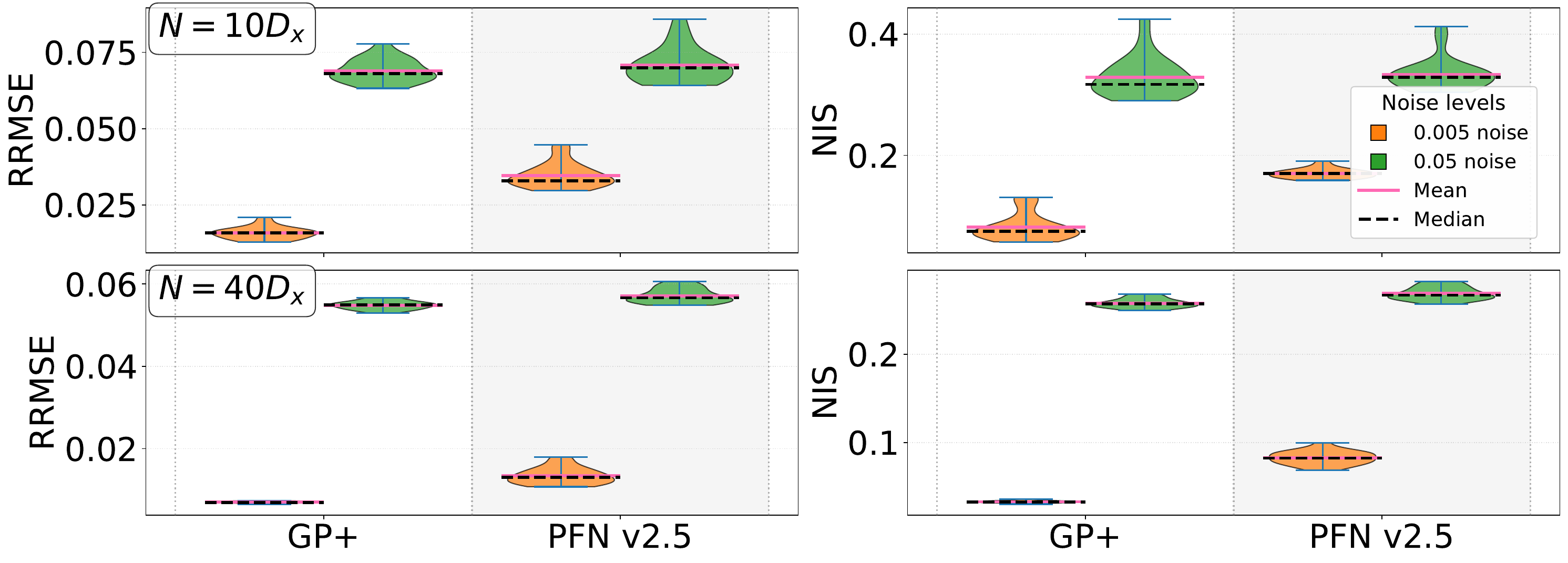}
        \subcaption{Wing Weight: D$_x$=10}
        \label{fig results wing}
    \end{minipage}
    \hfill
    \begin{minipage}[b]{0.48\textwidth}
        \centering
        \includegraphics[width=\textwidth]{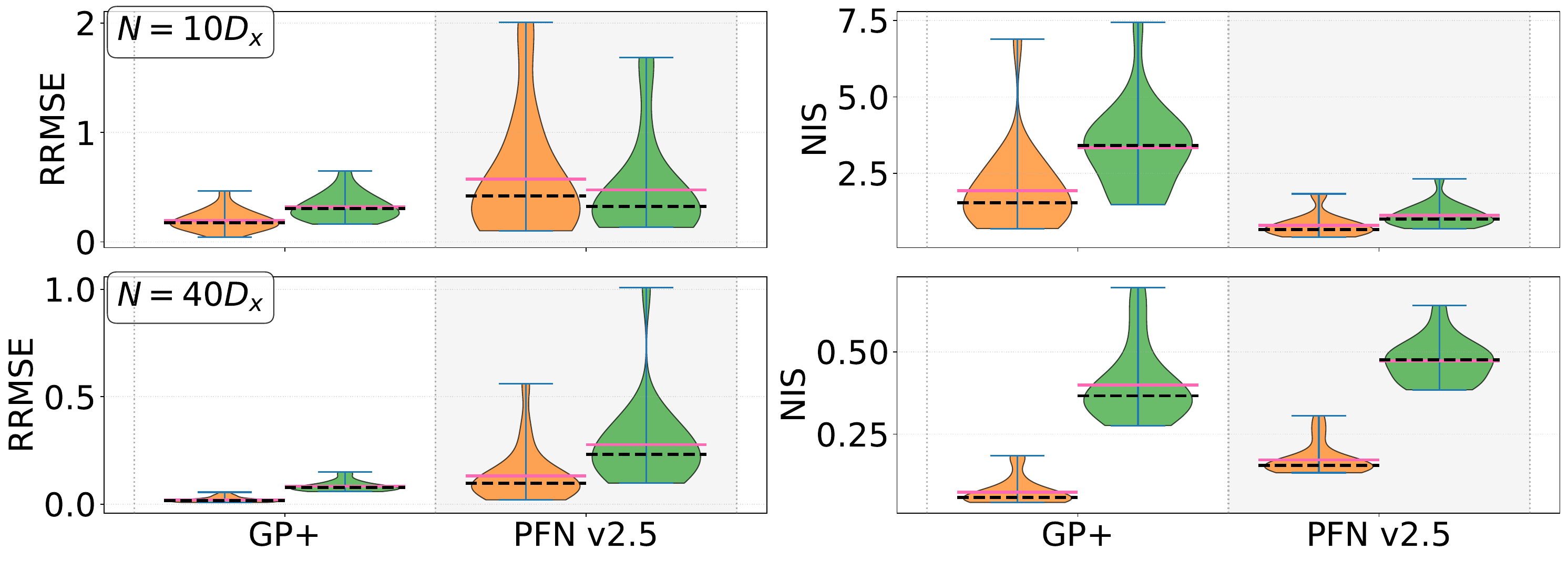}
        \subcaption{Buckling: D$_x$=4}
        \label{fig results buckling}
    \end{minipage}
    
    \vspace{1em}
    
    \begin{minipage}[b]{0.48\textwidth}
        \centering
        \includegraphics[width=\textwidth]{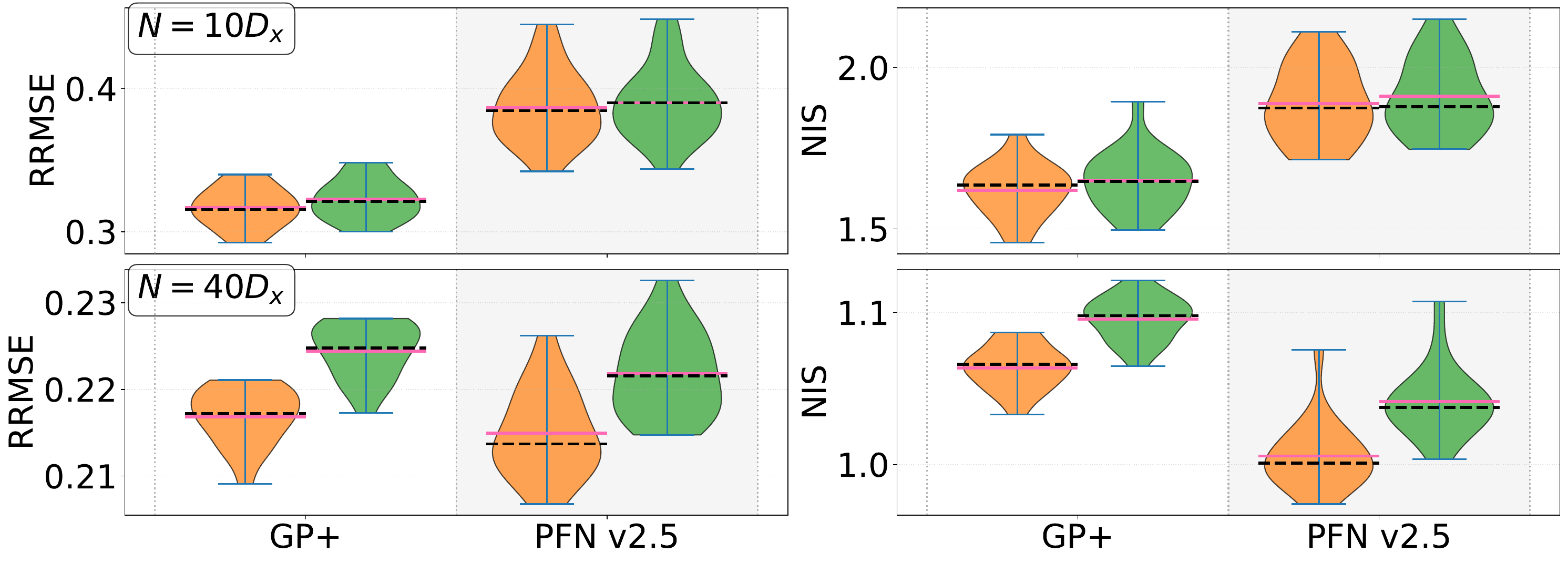}
        \subcaption{Ackley: D$_x$=20}
        \label{fig results ackley}
    \end{minipage}
    \hfill
    \begin{minipage}[b]{0.48\textwidth}
        \centering
        \includegraphics[width=\textwidth]{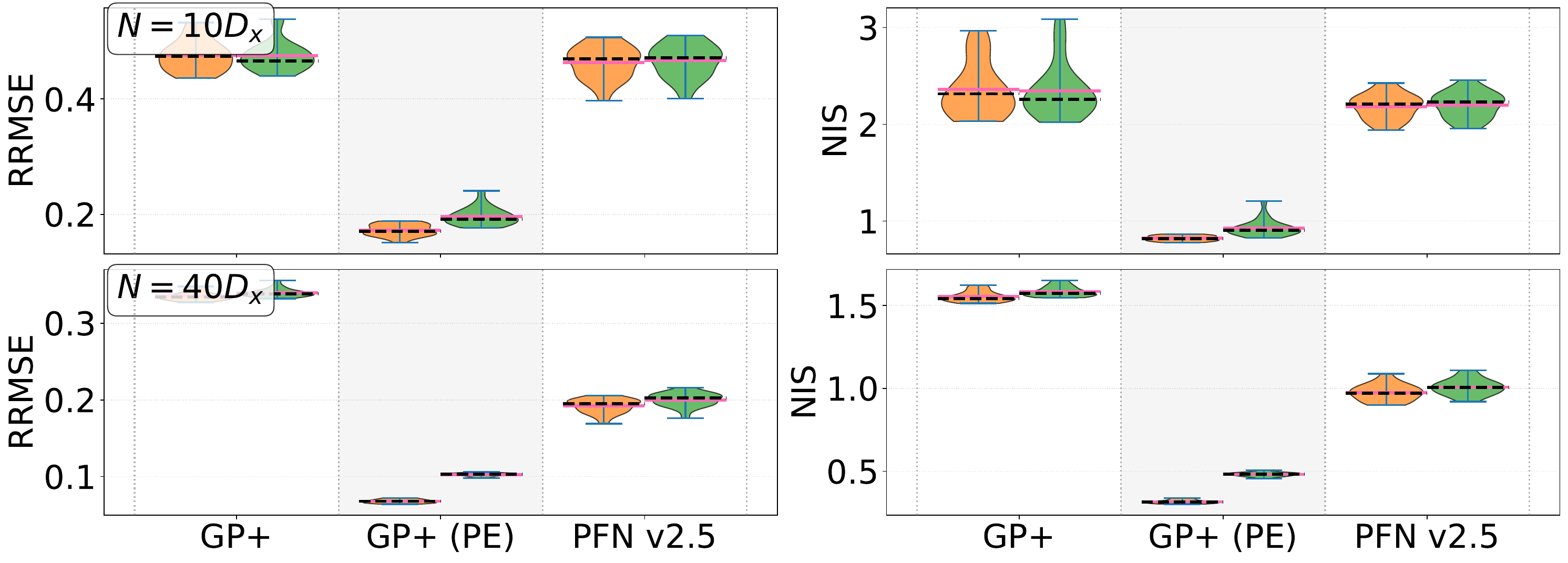}
        \subcaption{Dixon-Price: D$_x$=20}
        \label{fig results dixon}
    \end{minipage}
    
    \vspace{1em}
    
    \begin{minipage}[b]{0.48\textwidth}
        \centering
        \includegraphics[width=\textwidth]{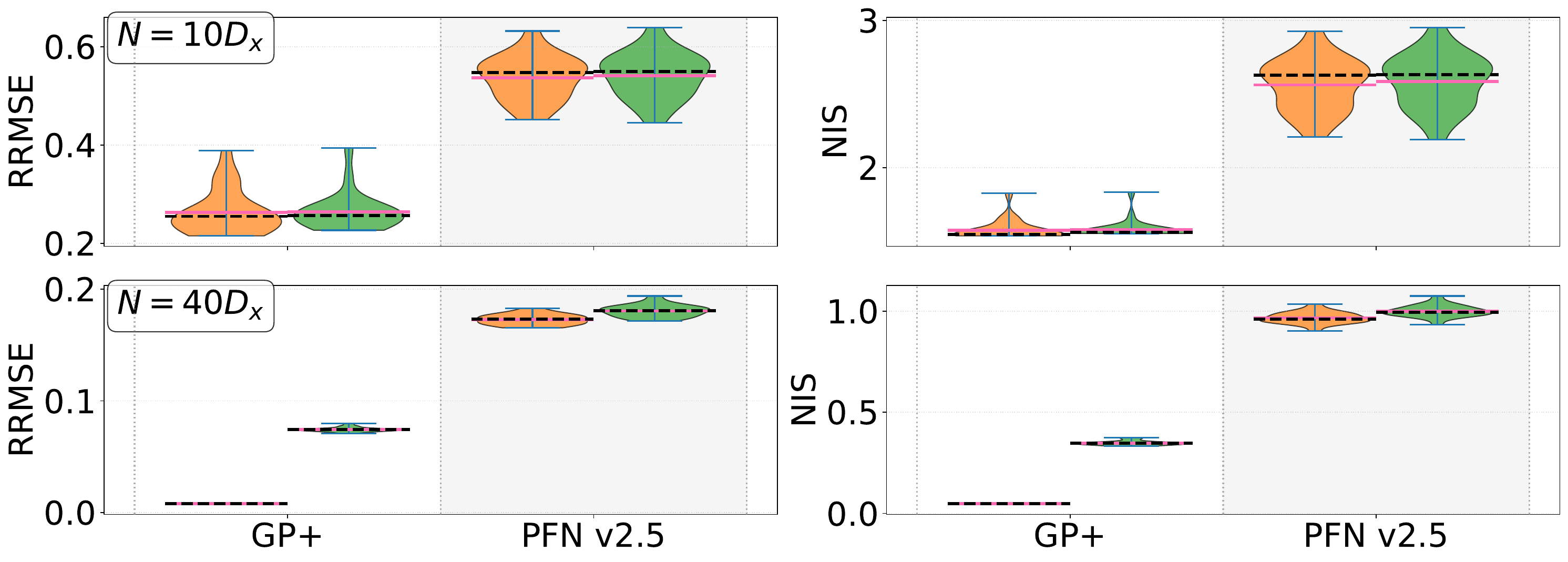}
        \subcaption{Griewank: D$_x$=40}
        \label{fig results griewank}
    \end{minipage}
    \hfill
    \begin{minipage}[b]{0.48\textwidth}
        \centering
        \includegraphics[width=\textwidth]{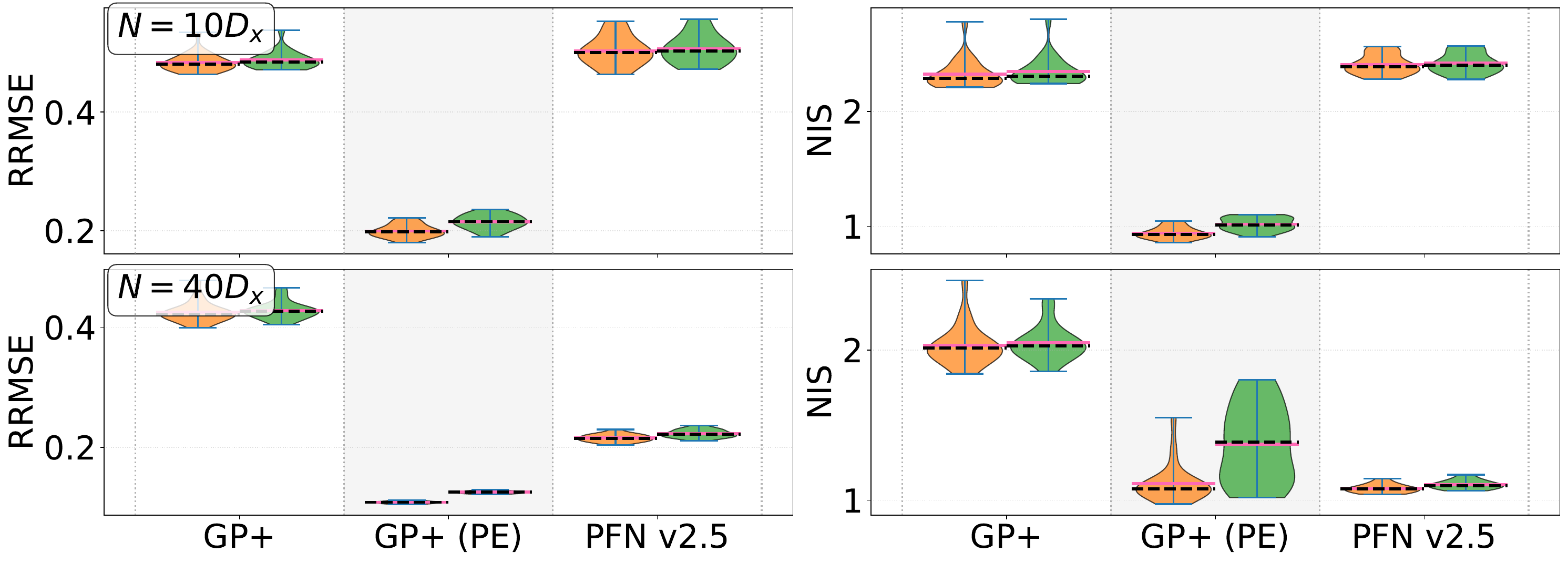}
        \subcaption{Rosenbrock: D$_x$=40}
        \label{fig results rosenbrock}
    \end{minipage}
    
    \caption{\textbf{Visualization of a subset of simulations listed in \Cref{tab results}:} RRMSE and NIS are used to compare TabPFN and GP across six benchmarks whose input dimensionality, $D_x$, ranges from $4$ to $40$ (only Buckling includes categorical inputs). Size of the training data is either $10D_x$ or $40D_x$ and noise is either $\varepsilon \sim \mathcal{N}(0, (0.005\times c)^2)$ or $\varepsilon \sim \mathcal{N}(0, (0.05\times c)^2)$. For Dixon-Price and Rosenbrock, we also consider the power exponential (PE) kernel to show the effect of kernel type on GPs. All the GPs are trained with the \textit{default} settings of GP+ package \cite{RN2079}. GPs are used on CPU while TabPFN is used on GPU. }

    \label{fig results}
\end{figure*}

This outcome seems counterintuitive at first as one may expect a ``trained'' model to outperform but as explained below it is primarily due to the fact that the GPs in \Cref{tab results} are trained via default settings of GP+ (w.g., Gaussian kernel). For example, in the Wing Weight benchmark in \Cref{fig results wing}, GP+ achieves very high emulation accuracy where the test RRMSE approaches the injected noise levels (e.g., $0.005$ and $0.05$). GP+ achieves lower RRMSE on Wing Weight across all reported settings in \Cref{tab results}, and generally lower NIS, with TabPFN marginally outperforming GP+ on NIS only in the higher-noise, small-data setting. In the lower-noise setting, TabPFN’s RRMSE and NIS distributions remain shifted upward relative to GP+, suggesting it does not fully exploit the near-deterministic regime as effectively as the GP baseline. Another notable behavior in this regime is that TabPFN’s NIS distribution remains relatively tight across runs even when its RRMSE varies substantially. This connotes that TabPFN’s predictive intervals remain fairly well-behaved across splits, but its predictive mean can be less accurate than the uncertainty score would imply.

In the Buckling example, TabPFN v2.5 exhibits considerably higher variability in RRMSE than GP+ across all reported settings, reflecting sensitivity of its mean predictions to the specific training split when context is very small. Despite unreliable mean predictions, TabPFN produces substantially better-calibrated intervals than GP+ at $N=10D_x$. As $N$ increases to $40D_x$, GP+ improves dramatically in both RRMSE and NIS, while TabPFN's gains are comparatively modest. In \Cref{fig results buckling}, TabPFN's RRMSE remains highly variable across runs while its NIS stays comparatively tight and well-behaved, suggesting TabPFN maintains appropriate predictive uncertainty even in runs where its median predictions are poor. Testing at higher data volumes (results not shown) confirmed that GP+ improves more efficiently with additional data than TabPFN, whose accuracy stagnates in this case.


Ackley is a highly multimodal benchmark for which both methods achieve relatively high RRMSEs and NISs at $N = 10D_x$, with TabPFN typically exhibiting greater run-to-run variability and higher medians than GP+. When the training set increases to $N = 40D_x$, both models improve and the two methods become comparable: TabPFN often achieves slightly lower median RRMSE and NIS, while GP+ tends to retain tighter distributions across runs, see \Cref{fig results ackley} for a subset of results. We note that our studies are limited to $N \leq 40D_x$ for high-dimensional problems as we refrain from using scalable GP methods. Comparison of TabPFN to scalable GPs is left for future work.


The Griewank benchmark highlights a clear advantage for GP+. In the low-data regime ($N=10D_x$), TabPFN v2.5 exhibits substantially higher RRMSE and NIS, indicating difficulty learning the underlying structure of the objective from limited context. As the dataset size increases to $N=40D_x$, GP+ improves sharply and reaches very low RRMSE and NIS, while TabPFN improves only modestly and remains shifted upward in both metrics. Griewank illustrates a regime where the GP prior and default training procedure yield strong sample efficiency and stable uncertainty estimates, whereas TabPFN requires substantially more data to become competitive (see \Cref{tab results}).

We observe interesting trends in Dixon-Price and Rosenbrock benchmarks where GP+ outperforms or is comparable to TabPFN in low-data high-dimensional regimes (e.g., $N=10D_x$ and $D_x = 40$) but loses advantage as $N$ increases to the extent that, unlike TabPFN, little gains are obtained in both RRMSE and NIS. As discussed below, these counterintuitive trends are due to the fixed GP+ defaults used across all problems. 




    

\subsection{Effect of Kernel and Mean Choice on GP}
There are two cases shown in \Cref{fig results dixon,fig results rosenbrock} where, unlike TabPFN, the performance of GP+ does not improve noticeably as more samples are used in training. Normally, one leverages the information gained from the GP's interpretable hyperparameters or loss function value (i.e., log marginal likelihood) to implement different strategies that can enhance the models. These techniques increase the cost and complexity of training GPs, which TabPFN users do not typically have to use.

We show in \Cref{fig results dixon,fig results rosenbrock} that simply changing the kernel from Gaussian to PE significantly improves the performance of the GP. It is important to note that this adds another hyperparameter and as a result increases the training cost. 
Additionally, when training GPs on high-dimensional (e.g., $D_x=40$) versions of the Rosenbrock function, we observed that the estimate of the constant mean depended substantially on the optimization initialization. Based on this observation we experimented with a zero-mean GP which performed much better. Since we do not use scalable GP approaches, we reduced the number of initializations during hyperparameter optimization from the default $16$ to $8$ which reduced training costs, but may have contributed to a higher overall NIS.
This approach indicates that by paying the price of fine-tuning the model setup and training for a particular application, GPs can provide highly accurate results. 


\subsection{Computational Cost}

\Cref{tab results} reports the combined training and prediction time for GP+ and end-to-end inference time (conditioning plus prediction) for TabPFN v2.5, where for GP+ the majority of the cost is spent during training, while for TabPFN it is spent during prediction. These timings are hardware dependent because we run the GPs on CPU and TabPFN on GPU, so they should be interpreted as representative wall-clock behavior rather than a device-normalized comparison. For reference, \Cref{tab results} also reports TabPFN's wall-clock time on CPU (in parentheses), which is substantially higher than its GPU time and highlights the importance of dedicated hardware for transformer-based inference.

GP+ training time is not solely a function of $N$ and $D_x$; it also depends on how easily the log-marginal-likelihood landscape can be optimized for a given problem. For example, on Wing Weight with $D_x = 10$ and $N = 10D_x$, GP+ completes training and prediction in roughly 1.1 s, whereas Buckling at the same training-size scaling takes 5--10 s depending on the noise level despite having fewer total training samples ($N = 40$ vs.\ $N = 100$), with the low-noise setting being slower due to a flatter likelihood surface that is harder to optimize. Despite this, TabPFN's end-to-end inference cost is roughly constant at $\approx 1.4$ s on GPU across these cases. As $N$ and $D_x$ grow, exact GP training quickly dominates: on Griewank with $D_x = 40$ and $N = 40D_x$, GP+ takes $\approx 8.3 \times 10^2$ s, whereas TabPFN finishes in $\approx 4.8$ s on GPU. This pattern is consistent across the high-dimensional $N = 40D_x$ cases.


A key detail is that our reported prediction times are measured on a large test set ($N_\text{test} = 5000$). If only a small number of queries were needed per fit (e.g., under 100), TabPFN's end-to-end inference time would drop noticeably since the test set substantially outsizes the training context ($N_c \leq 40D_x$) in our setup.


These trade-offs matter most in downstream use. In Bayesian optimization, we often fit once and score many candidate points, so fast post-fit GP prediction is an advantage when training remains affordable. Conversely, when the dataset is updated frequently and only a small number of predictions are needed per update, TabPFN can be attractive because it avoids per-dataset hyperparameter optimization.


\subsection{Additional Studies and Practical Implications}
To assess robustness beyond the synthetic Gaussian-noise setting, we additionally evaluated both methods under Student's $t$-distributed noise ($\nu = 4$) and on two real-world tabular regression datasets (Elevators and Pumadyn). The qualitative trends reported above were preserved in both cases, suggesting our conclusions are not an artifact of the Gaussian observation model that favors GPs. Full tables and figures are provided in our GitHub repository.

In \Cref{tab results real}, we see that, with respect to their median performance, TabPFN outperforms GP+ in the Elevators problem and the opposite is observed in the Pumadyn problem. It is important to note that the performance of both models does not improve much as the training dataset increases in size, and the gap between the median performance remains consistent as more data is added to the training set. 
Similar to \Cref{tab results} we observe that the method with the best RRMSE also provides the best NIS; indicating that both methods provide decent UQ. 

Across the benchmarks, GP+ with a single default configuration is often competitive with TabPFN v2.5, but the performance depends on how well the GP prior matches the function class and on problem difficulty (e.g., high dimensionality and strong multimodality). In very low-noise settings, GP+ more reliably tracks the noise floor, while TabPFN tends to retain a residual error even as more data are provided. TabPFN seems less suitable for high-precision applications that require errors near the simulator noise floor. Conversely, when the default GP kernel/mean is misspecified, TabPFN can narrow the gap or surpass GP+ as context grows. Practically, this suggests using GP+ defaults as a strong baseline and applying minor adjustments to the kernel/mean when performance saturates, before resorting to heavier tuning.

\begin{table*}[!h]
    \centering
    {\extracolsep{-10pt}
    \setlength{\tabcolsep}{5pt}
    \renewcommand{\arraystretch}{1.15}
    \scriptsize
    \caption{\textbf{Summary of results for real-world datasets:} RRMSE, NIS, and wall-clock time (medians across 20 runs, see GitHub for statistics). TabPFN reports GPU time with CPU time in parentheses. GP runs on CPU; TabPFN on GPU. All GPs use default GP+ settings.}
    \begin{tabular}{>{\centering\arraybackslash}p{1.2cm}>{\centering\arraybackslash}p{0.5cm}>{\centering\arraybackslash}p{0.3cm}|ccc|ccc}
    \hline
    \multicolumn{3}{c|}{} & \multicolumn{3}{c|}{\textbf{GP+}} & \multicolumn{3}{c}{\textbf{TabPFN v2.5}} \\
    \cline{4-9}
    \textbf{Problem} & \textbf{N} & \textbf{D$_x$} & \textbf{RRMSE} & \textbf{NIS ($A{+}B$)} & \textbf{CPU Time (s)} & \textbf{RRMSE} & \textbf{NIS ($A{+}B$)} & \textbf{GPU Time (s)} \\
    \hline
    \textbf{Elevators} & 10D$_x$ & 18 & 0.381 & 2.060 (1.292 + 0.768) & 2.40 & \textbf{0.336} & \textbf{1.591} (1.274 + 0.317) & \textbf{0.97} \\[2pt]
     & 20D$_x$ & 18 & 0.347 & 1.827 (1.275 + 0.552) & 9.43 & \textbf{0.302} & \textbf{1.413} (1.107 + 0.306) & \textbf{1.06} \\[2pt]
     & 40D$_x$ & 18 & 0.324 & 1.685 (1.245 + 0.413) & 43.62 & \textbf{0.288} & \textbf{1.327} (1.029 + 0.297) & \textbf{1.23} \\[2pt]
    \hline
    \textbf{Pumadyn} & 10D$_x$ & 32 & \textbf{0.206} & \textbf{0.971} (0.762 + 0.209) & 13.69 & 0.257 & 1.202 (0.952 + 0.250) & \textbf{1.60} \\[2pt]
     & 20D$_x$ & 32 & \textbf{0.194} & \textbf{0.900} (0.736 + 0.164) & 71.44 & 0.252 & 1.162 (0.919 + 0.243) & \textbf{1.89} \\[2pt]
     & 40D$_x$ & 32 & \textbf{0.186} & \textbf{0.950} (0.865 + 0.085) & 423.25 & 0.250 & 1.137 (0.907 + 0.231) & \textbf{2.47} \\[2pt]
    \hline
    \end{tabular}
    \label{tab results real}
    }
\end{table*}
    \section {Conclusion and Future Directions} \label{sec conclusion}

We compared the performance of tabular PFNs and GPs on a set of benchmarks. TabPFN offers a computationally cheap alternative to traditional emulators that is able to generalize to many datasets in high-dimensional, large-data, and mixed-input regimes. It is becoming an increasingly popular method due to its ability to reduce the computational overhead required from fitting a good emulator. This FM can remove the need to manually preprocess a dataset or train and tune a model. Our evaluation of GPs and TabPFN showed that TabPFN is a cost-effective method to obtain accurate point estimates, but its ability does not tend to rival GPs if they are equipped with a good kernel.

We observed that increasing the dataset size in some of the benchmarks (i.e., Dixon-Price and Rosenbrock) negligibly improved performance while dramatically increasing the training costs of GPs. 
Meanwhile, TabPFN was able to leverage more training samples to approximate the Bayesian posterior predictive distribution better and this shows its ability to scale well with larger datasets. This trend is due to the fact that all our GPs were trained with default settings of GP+ while it is common practice to fine-tune the mean function or kernel of the GP slightly based on the application. As shown in our results, fine-tuning can substantially improve the performance of GPs but does require inspecting the hyperparameters and the loss function. In contrast, our attempts to tune TabPFN beyond its defaults did not yield meaningful improvements, indicating that the default configuration is already a strong operating point and that the user effort saved on the TabPFN side is unlikely to be recovered through tuning. 

From an uncertainty perspective, both methods can produce reasonable intervals, but fail differently and offer different levels of interpretability. GP behavior can often be diagnosed through learned hyperparameters (e.g., noise and lengthscales), and GP+ exposes a direct decomposition of uncertainty: the learned likelihood noise (nugget) corresponds to aleatoric uncertainty, while the posterior variance of the latent function reflects epistemic uncertainty. TabPFN also outputs a flexible predictive distribution, but its uncertainty is produced by a neural network and is not explicitly parameterized into observation-noise and latent-function components.

Our studies excluded some recent advances on tabular FMs and on GPs including kernel construction and log-marginal likelihood optimization in high-dimensional, complex, or large-data settings \cite{negarandeh2026non,RN2201,binois2022survey}. Also, the complexity of our benchmarks was limited to two real-world datasets and six analytic problems with Gaussian or Students' t-distribution noise. Additionally, while RRMSE and NIS are useful metrics for assessing emulation performance, we believe more insights can be obtained via application-oriented comparisons such as Bayesian optimization, multi-fidelity modeling, or multi-task emulation. 
These directions will be pursued in our future research.

    \section*{Acknowledgments}
We appreciate the support from the Office of the Naval Research (award number N000142312485) and National Science Foundation (award number 2238038).
    \bibliography{R_Ref}
    
    
    
\end{document}